\documentclass{article}
\usepackage{spconf,amsmath,epsfig}
\usepackage{graphicx}
\usepackage{multirow}
\usepackage{booktabs}
\usepackage{amssymb}
\usepackage{setspace}

\usepackage[linesnumbered, ruled]{algorithm2e}
\SetKwRepeat{Do}{do}{while}

\let\OLDthebibliography\thebibliography
\renewcommand\thebibliography[1]{
  \OLDthebibliography{#1}
  \setlength{\parskip}{0pt}
  \setlength{\itemsep}{0pt plus 0.3ex}
}
\graphicspath{{figures/}}

\begin{document}\sloppy

\title{Camera-aware Style Separation and Contrastive Learning for Unsupervised Person Re-identification}
\name{Xue Li, Tengfei Liang, Yi Jin{\rm\textsuperscript{*}}, Tao Wang, Yidong Li \thanks{*Corresponding author.}}
\address{School of Computer and Information Technology, Beijing JiaoTong University, Beijing, China \\ \{xue.li, tengfei.liang, yjin, twang, ydli\}@bjtu.edu.cn}
\maketitle

\begin{abstract}
Unsupervised person re-identification (ReID) is a challenging task without data annotation to guide discriminative learning.
Existing methods attempt to solve this problem by clustering extracted embeddings to generate pseudo labels.
However, most methods ignore the intra-class gap caused by camera style variance, and some methods are relatively complex and indirect although they try to solve the negative impact of the camera style on feature distribution.
To solve this problem, we propose a camera-aware style separation and contrastive learning method (CA-UReID), which directly separates camera styles in the feature space with the designed camera-aware attention module.
It can explicitly divide the learnable feature into camera-specific and camera-agnostic parts, reducing the influence of different cameras.
Moreover, to further narrow the gap across cameras, we design a camera-aware contrastive center loss to learn more discriminative embedding for each identity.
Extensive experiments demonstrate the superiority of our method over the state-of-the-art methods on the unsupervised person ReID task.
\end{abstract}


\section{Introduction}

\begin{figure}[t]
\centering
\includegraphics[width=\linewidth]{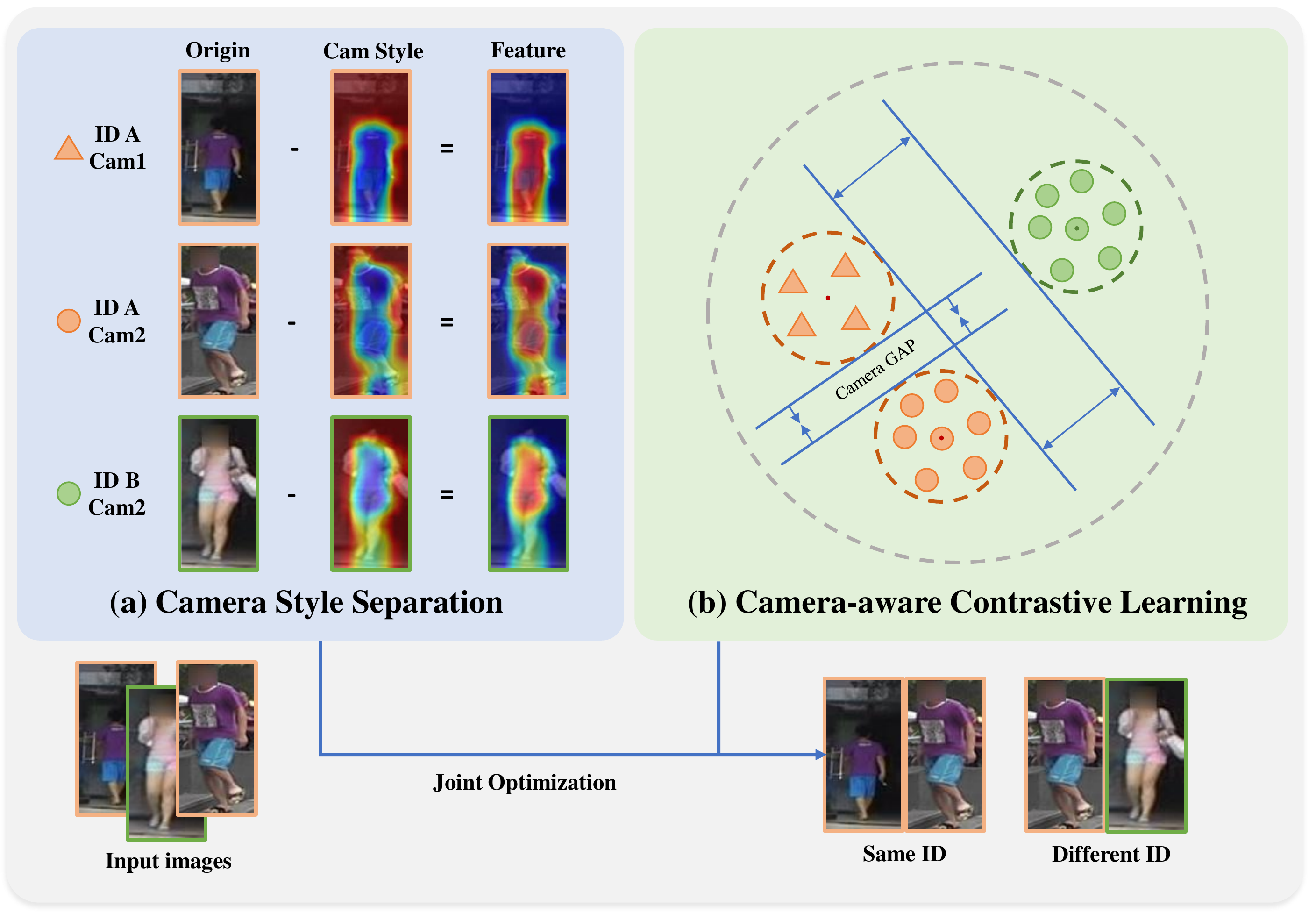}
\caption{A high-level overview of our proposed method: the CA-UReID method disentangles camera-agnostic feature from the original feature in a explicit way. By jointly optimized with the camera-aware contrastive learning, our method can effectively reduce the intra-class distance and increase the inter-class distance with high matching accuracy.}
\label{fig_conception}
\vspace{-10pt}
\end{figure}

Person re-identification (ReID) aims at retrieving images of the same person across cameras with non-overlapping views, which has important applications in the field of intelligent surveillance and public security. 
In the past few years, researchers pay more and more attention to this task and make great research progress.
However, existing high-performance methods mainly use the supervised learning framework with identities' annotation and the process of building a dataset is time-consuming and laborious.
Besides, the supervised models need to be retrained in a new scenario, which have poor scalability and are not conducive to the practical deployment.
Therefore, some researchers recently turn their attention to unsupervised person ReID (UReID)~\cite{DBLP:conf/aaai/HuangPJLX20, DBLP:conf/nips/Ge0C0L20,DBLP:journals/corr/abs-2103-16364, DBLP:conf/aaai/0001LHG021}, which has great potential value to get rid of dependence on manual annotation and become an emerging research hotspot.

Existing unsupervised person ReID methods are mainly divided into two categories: unsupervised domain adaptive (UDA) methods and purely unsupervised learning (USL) methods.
UDA methods~\cite{DBLP:conf/aaai/HuangPJLX20,DBLP:conf/iclr/GeCL20,DBLP:conf/nips/Ge0C0L20,DBLP:conf/aaai/ZhengLZZZ21} first train the model on a supervised source-domain dataset to make the model have a certain discriminant ability for identities.
Then the training is carried out in the target domain without annotation to transfer the ability.
However, the performance of UDA methods is greatly affected by the quality and size of the annotated source domain datasets, which constrains their scalability.
In contrast, the purely unsupervised methods~\cite{DBLP:conf/aaai/0001LHG021,DBLP:journals/corr/abs-2103-16364} that requires only the unlabeled target dataset is more flexible. 
USL methods mainly rely on clustering to generate pseudo labels for supervised learning in the target dataset with the strategy of contrastive learning.
However, most USL methods~\cite{DBLP:conf/eccv/LiZG18, DBLP:conf/bmvc/ChenZG18, DBLP:conf/iccv/WuLYLLL19} only focus on improving the differences between classes and ignore the intra-class differences caused by camera style variance, including camera property and environment information of cameras, etc.
Several methods~\cite{DBLP:journals/tist/TianTLTZF21,DBLP:conf/cvpr/XuanZ21, DBLP:conf/aaai/0001LHG021} that consider the influence of cameras are complicated and indirect.

Based on the above observations, we propose a novel purely unsupervised learning method, the camera-aware style separation and contrast learning network (CA-UReID), which directly disentangles camera style at the level of feature map and utilize camera-aware contrastive learning to learn more discriminative feature for each identity (Fig.\ref{fig_conception}).
Specifically, we first design a camera style separation (CSS) module that divides the extracted features into camera-specific and camera-agnostic parts with attention mechanism, so as to alleviate the impact of camera style variance on performance.
In the camera-specific branch, the disentangled attention module captures the camera style information under the guidance of the proposed camera separation loss.
Then, by using complementary attention maps, the camera-agnostic branch can enforce the network to focus on the persons, which extracts more camera-invariant features with the compact intra-class distribution.  
After the CSS module, we propose a camera-aware contrastive center (CACC) loss into the optimization constraints to further narrow the gap between cameras within the class and increase the feature distance of different classes.
Moreover, our CACC loss considers the constraints on the sample set under the same camera rather than individual samples, which is verified by us as a more stable and efficient way and is not easy to be affected by noise samples.
Through joint optimization with the designed camera style separation module and camera-aware contrastive center loss, our method can learn more robust features of identities for the final retrieval.

\vspace{2pt}
Our main contributions are summarized as follows:
\vspace{-5pt}
\begin{itemize}
  \item We propose a novel camera style separation module to explicitly disentangle camera-specific information from feature maps and reduce intra-class variance.
  
  \vspace{-5pt}
  \item We design a more robust camera-aware contrastive center loss, which further enhances camera-agnostic features at center level to get discriminative embeddings.
  
  \vspace{-5pt}
  \item Extensive experiments on mainstream datasets (Market1501~\cite{DBLP:conf/iccv/ZhengSTWWT15} and DukeMTMC-reID~\cite{DBLP:conf/eccv/RistaniSZCT16}) show superior performance of our proposed CA-UReID method.
\end{itemize}

\section{Related Work}

\textbf{Unsupervised Person Re-identification.}
Unsupervised person ReID methods can be categorized into the unsupervised domain adaptive (UDA) methods and the purely unsupervised learning (USL) methods.
The main difference between them is whether to use labeled source domain datasets.
As for the UDA methods, some of them use GAN~\cite{DBLP:conf/cvpr/Deng0YK0J18,DBLP:conf/cvpr/WeiZ0018} for style transfer to narrow the gap between the source domain and target domain.
Some other methods are designed to generate more reliable pseudo labels: 
SpCL~\cite{DBLP:conf/nips/Ge0C0L20} introduces a cluster reliability criterion to retain reliable clusters and disassemble unreliable clusters to outlier instances.
DAAM~\cite{DBLP:conf/aaai/HuangPJLX20} assigns different weights to samples based on their distance from the class centers. 
UNRN~\cite{DBLP:conf/aaai/ZhengLZZZ21} measured the reliability of pseudo labels by inconsistencies between the teacher network and student network.
Compared with the UDA method, the USL method is more flexible and promising without manual annotations.
Some approaches explore special clustering methods, such as k-NN~\cite{DBLP:conf/eccv/LiZG18,DBLP:conf/bmvc/ChenZG18} and graphs~\cite{DBLP:conf/iccv/WuLYLLL19}, to generate pseudo-labels. 
MMCL~\cite{DBLP:conf/cvpr/WangZ20a} uses a memory-based multi-label classification loss to update the model. 
CycAs~\cite{DBLP:conf/eccv/WangZZLSLW20} learns pedestrian embedding from original videos to mine temporal information.
Most USL methods focus on the distance between classes but often ignore the intra-class variance caused by inconsistent camera styles. 
In contrast, our USL-based CA-UReID method considers the effects of camera style, which effectively improves the consistency within the class.

\vspace{4pt}
\noindent\textbf{Research of Camera Style in UReID.}
Eliminating the effects of camera style has recently been highlighted by a few unsupervised ReID methods.
CIDC~\cite{DBLP:journals/tist/TianTLTZF21} method transfers samples to each camera via StarGAN~\cite{DBLP:conf/cvpr/ChoiCKH0C18}.
Some other methods use intra-camera and inter-camera combined training. 
In the intra-camera training stage, a classifier is set under each camera, and all the classifiers share a backbone network. 
In the inter-camera training stage, existing methods design different strategies to unify the knowledge learned from different cameras:
IICS~\cite{DBLP:conf/cvpr/XuanZ21} uses a combination of Instance Normalization (IN) and Batch Normalization (BN) to improve the generalization ability of classifiers. 
CAP~\cite{DBLP:conf/aaai/0001LHG021} designs camera-based clusters to pull positive clusters and push nearest negative proxies. 
MetaCam~\cite{DBLP:conf/cvpr/YangZLCLLS21} uses meta-learning to learn robust models for cameras. 
Most of these methods have complex training processes, and their way of eliminating the influence of camera style is indirect.
Different from them, our proposed CA-UReID method separates camera style in an explicit way by disentangling camera-specific information in the feature space. 
Moreover, compared with the existing CAP~\cite{DBLP:conf/aaai/0001LHG021} and ICE~\cite{DBLP:journals/corr/abs-2103-16364} methods, 
our designed camera-aware contrastive loss considers the overall distribution in each camera-aware cluster, which is more robust with better performance.

\section{Method}

\begin{figure*}[t]
\centering
\includegraphics[width=\linewidth]{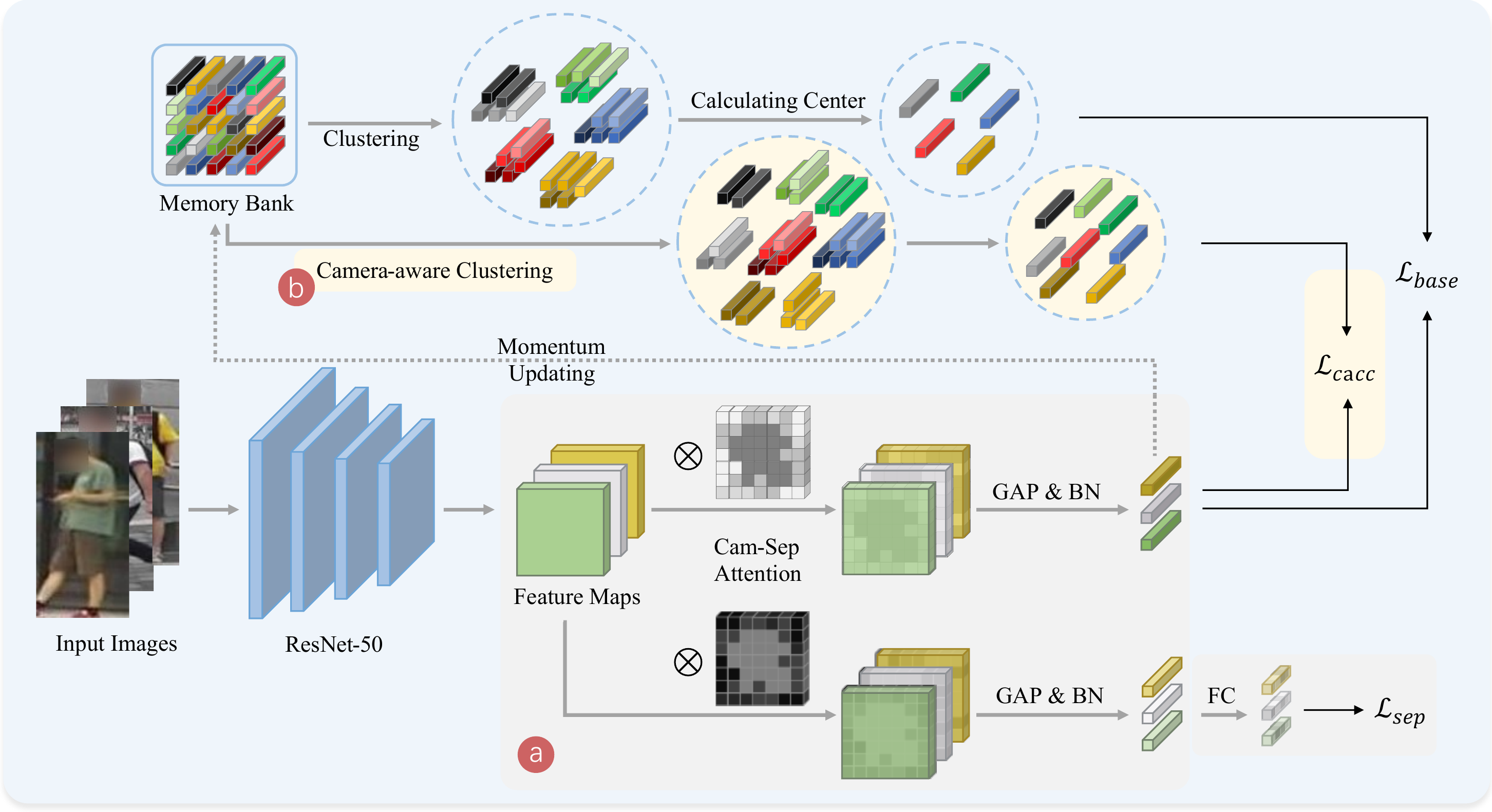}
\vspace{-18pt}
\caption{
The framework of our proposed method.
(a) The two-branch camera style separation module, including the camera-specific branch and the camera-agnostic branch.
(b) The pipeline of calculating the camera-aware contrastive center loss.
}
\vspace{-10pt}
\label{fig_overall_structure}
\end{figure*}

In this section, we introduce our camera-aware style separation and contrastive learning (CA-UReID) method in detail.
In Sec.~\ref{subsec_overview}, we first describe the overall architecture of CA-UReID, including the structure of network and the overall objective function during the training.
Then we introduce our proposed camera style separation module and camera-aware contrastive center loss in the next two subsections.

\subsection{Overall Structure} \label{subsec_overview}
As shown in Fig.~\ref{fig_overall_structure}, our method follows the memory bank-based framework, constructing an unsupervised learning network.
The overall process of CA-UReID mainly includes three stages: identities' feature extraction, two-branch camera style separation, and self-supervised contrastive learning.

In the feature extraction stage, let $\mathcal{X}=\{x_1,x_2,…,x_N\}$ denote the person ReID dataset, and $N$ is the number of samples.
Specifically, given an input sample $x_i$, we use ResNet-50 as the backbone to extract the corresponding feature map $F_{x_i}\in \mathbb{R}^{H \times W \times C}$, 
where $H$, $W$, $C$ denote height, width and channel number respectively. 
In the stage of two-branch camera style separation, the designed camera separation (Cam-Sep) attention module is used to disentangle the camera-specific feature map $F_{x_i}^{sp}$ and camera-agnostic feature map $F_{x_i}^{ag}$ from $F_{x_i}$.
The $F_{x_i}^{sp}$ is fed into the camera-specific branch under the constraints of the proposed camera separation loss $\mathcal{L}_{sep}$, and we use the $F_{x_i}^{ag}$ to generate the camera-agnostic 2048-dim feature embedding $f_i$ with the global average pooling (GAP) and batch normalization (BN) operations.

During the last stage of self-supervised contrastive learning, as shown in the upper part of Fig.~\ref{fig_overall_structure}, our method follows the framework based on memory bank like some previous methods~\cite{DBLP:conf/nips/Ge0C0L20}.
Before training, the memory bank is initialized with the embeddings extracted from the whole training set by using ImageNet-pretrained ResNet-50.
At the beginning of each epoch, the k-reciprocal-coded Jaccard distance matrix of all pedestrian features in the bank is used for DBSCAN clustering to obtain pseudo labels $\mathcal{Y}=\{y_1,y_2…y_N\}$.
Base on the pseudo labels, our method can calculate the common InfoNCE loss~\cite{DBLP:journals/corr/abs-1807-03748} $\mathcal{L}_{base}$ and our proposed camera-aware contrastive center loss $\mathcal{L}_{cacc}$.
Let $\mathcal{M}$ ($\mathcal{M} \in \mathbb{R}^{d\times N}$) denotes the memory bank, where $d$ is the dimension of the embeddings.
At the end of each training iteration, extracted embeddings $f_i$ are used to update the memory bank $\mathcal{M}$ at the instance level with the momentum mechanism:

\vspace{-20pt}
\begin{eqnarray}
  \mathcal{M}[i]\leftarrow \mu \cdot \mathcal{M}[i]+\left(1- \mu \right) \cdot f_{i}~,
\end{eqnarray}
\vspace{-19pt}

\noindent where $\mathcal{M}[i]$ is the $i$-th item in the memory bank, storing the updated feature of sample $x_i$, and $\mu$ is a momentum coefficient that controls the updating speed.

In the training mode, the overall objective function of this CA-UReID method can be defined as follows:

\vspace{-20pt}
\begin{eqnarray}
  \mathcal{L}_{total}=\mathcal{L}_{base}+\lambda_{1} \cdot \mathcal{L}_{sep}+\lambda_{2} \cdot \mathcal{L}_{cacc}~,
  \label{eq_total_loss}
\end{eqnarray}
\vspace{-19pt}

\noindent  where $\lambda_{1}$ and $\lambda_{2}$ are hyper-parameters to balance these different components.
In the test mode, the $f_{i}$ extracted from image is the embedding used for final matching.




\vspace{-10pt}
\subsection{Camera Style Separation Module} \label{subsec_CSS}
\vspace{-5pt}

The camera style variance, including camera property and environment information, leads to the gap in intra-class features. 
To solve this problem, we propose the camera style separation (CSS) module to eliminate the camera style directly from the extracted feature maps.
Inspired by the way DAAM~\cite{DBLP:conf/aaai/HuangPJLX20} separates domain-related information in UDA framework, we design a novel strategy to disentangle camera-style information in our USL pipeline. 
As shown in Fig.~\ref{fig_overall_structure}a, the designed CSS module has two branches: the camera-specific branch and the camera-agnostic branch.

In the camera-specific branch, to separate camera-style features unrelated to pedestrians (e.g. background and camera property, etc.) from feature map, we reframe the learning of this branch as a camera classification problem.
Note that, although we don’t have ID annotation in USL mode, we can use the cameras annotations, which is easy to obtain in the actual collection of datasets.
Specifically, $F_{x_i}$ is fed into a Cam-Sep attention module to obtain a weight mask $Attn(F_{x_i})$ to focus on camera information. 
Here, we adopt the HA structure of HA-CNN~\cite{DBLP:conf/cvpr/LiZG18} as the attention module to obtain both spatial and channel attention weights.
Through element wise product with the feature map $F_{x_i}$, camera-specific feature map $F_{x_i}^{sp}$ is calculated by the following formula:

\vspace{-18pt}
\begin{eqnarray}
  F_{x_i}^{sp}=Attn(F_{x_i})\otimes F_{x_i}~.
\end{eqnarray}
\vspace{-15pt}

\noindent The camera-specific embedding $f_i^{sp}$ is generated by GAP and BN operation on $F_{x_i}^{sp}$.
To guide the attention of this branch, we design the camera separation loss $\mathcal{L}_{sep}$:

\vspace{-18pt}
\begin{eqnarray}
  \mathcal{L}_{sep}=\mathbb{E}[-y_i^{cam}log(\sigma(FC(f_i^{sp})))]~,
\end{eqnarray}
\vspace{-15pt}

\noindent where $y_i^{cam}$ is the camera label, $\sigma$ is the Softmax function, and $FC$ is the fully connected layer.
$f_i^{sp}$ is input into a classifier with the number of categories as the number of cameras to predict probability of belonging to each camera, which can enforce this branch to extract camera-related information.

In the camera-agnostic branch, we carry out element-wise products with the complementary mask of $Attn(F_{x_i})$ and 
the original feature map $F_{x_i}$ to obtain the camera-agnostic feature map $F_{x_i}^{ag}$, which is defined as follows:

\vspace{-18pt}
\begin{eqnarray}
  F_{x_i}^{ag}=(1-Attn(F_{x_i}))\otimes F_{x_i}~.
\end{eqnarray}
\vspace{-15pt}

\noindent Then the camera-agnostic feature vector $f_i$ is obtained from $F_{x_i}^{ag}$ and used to calculate $\mathcal{L}_{base}$ and $\mathcal{L}_{cacc}$.
By optimizing $\mathcal{L}_{sep}$, the influence of camera styles is reduced and the model can learn more robust pedestrian features.

\vspace{-10pt}
\subsection{Camera-aware Contrastive Center Loss} \label{subsec_CACC}
\vspace{-5pt}

To further enhance the network to extract camera-invariant embeddings $f_i$, we propose a novel camera-aware contrastive center (CACC) loss $\mathcal{L}_{cacc}$ to narrow the feature distance of different cameras in the same class and increase the feature distance of different classes.

\begin{figure}[t]
\centering
\includegraphics[width=\linewidth]{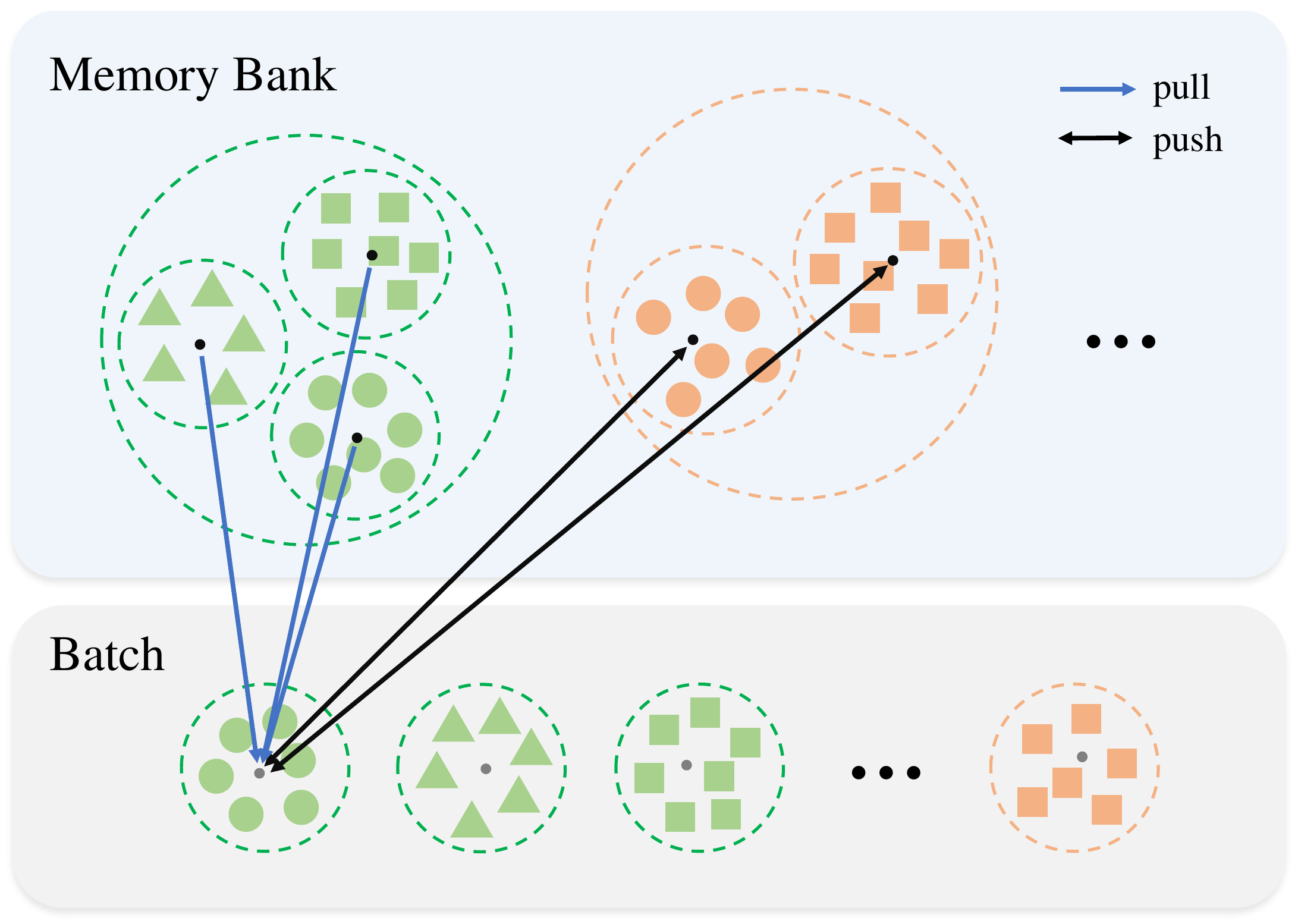}
\vspace{-18pt}
\caption{The camera-aware contrastive center loss. The geometrical patterns with different colors represent different ID's embeddings, and different shapes represent different cameras.}
\vspace{-10pt}
\label{fig_cacc_loss}
\end{figure}

As shown in Fig.~\ref{fig_overall_structure}b, different from the common contrastive learning loss $\mathcal{L}_{base}$ in UReID, the $\mathcal{L}_{cacc}$ considers the camera in the clustering process.
Besides, unlike other camera-aware contrastive losses (ICE~\cite{DBLP:journals/corr/abs-2103-16364}, CAP~\cite{DBLP:conf/aaai/0001LHG021}), our proposed $\mathcal{L}_{cacc}$ constrains the embeddings at the set level by using a camera center instead of a sample as an anchor (Fig.~\ref{fig_cacc_loss}), attempting to avoid the impact of noise samples.
Specifically, given the camera number $c$ and identity class $k$ for each embedding stored in the memory bank $\mathcal{M}$ and a batch, the camera center $g_k^{c}$ and $p_k^{c}$ can be respectively formulated as:

\vspace{-13pt}
\begin{eqnarray}
  g^{c}_k=\frac{1}{N^c_{k}}\sum_{\mathcal{M}[i] \in k \cap c} \mathcal{M}[i],~~~~p^{c}_k=\frac{1}{\hat{N}^c_{k}}\sum_{f_{i} \in k \cap c} f_{i}~,
\end{eqnarray}
\vspace{-10pt}

\noindent where $N^c_{k}$ is the number of all embeddings of identity class $k$ with camera number $c$ in $\mathcal{M}$, and $\hat{N}^c_{k}$ is the number of embeddings with class $k$ and camera number $c$ in a training batch.

In each batch, by using $p^{c}_k$ as an anchor and  
combining all positive camera centers $g_i$ and k-nearest negative camera centers $g_j$, 
our proposed $\mathcal{L}_{cacc}$ can be formulated as:

\vspace{-13pt}
\begin{eqnarray}
  \mathcal{L}_{cacc}=-\frac{1}{G}\sum_{i\in k}log \frac{S(p^{c}_k, g_i)}{S(p_k^{c} , g_i)+\sum_{j=1}^{J}S(p_k^{c} , g_j)}~,
\end{eqnarray}
\vspace{-10pt}

\noindent where $G$ is the number of $g_i$, $J$ is the number of $g_j$, $S(p_k^{c} , g_i)=exp(p_k^{c} \cdot g_i/\tau_{cacc})$, and $\tau_{cacc}$ is a temperature hyper-parameter.
By optimizing the $\mathcal{L}_{cacc}$ constrained on the center within each camera (shown in Fig.~\ref{fig_cacc_loss}), our method can effectively pull the intra-class features and push the inter-class features to obtain more discriminative embeddings.

\vspace{-5pt}
\section{Experiments}

\begin{table}[htbp]
  \centering
  \caption{Comparison with state-of-the-art methods.}
  \fontsize{8}{9.5}\selectfont
  \setlength{\tabcolsep}{0.4mm}
  \begin{tabular}{c|c|c|cc|c|cc}
  \hline
  \hline
  \multirow{2}[0]{*}{Method} & \multirow{2}[0]{*}{Venue} & \multirow{2}[0]{*}{Source} & \multicolumn{2}{c|}{Market} & \multirow{2}[0]{*}{Source} & \multicolumn{2}{c}{Duke} \\\cline{4-5}\cline{7-8}
        &       &       & mAP   & Rank1  &       & mAP   & Rank1  \\
  \hline
  \multicolumn{8}{l}{\textbf{Unsupervised Domain Adaptation}}    \\
  \hline
  MMCL~\cite{DBLP:conf/cvpr/WangZ20a}       & CVPR2020 & Duke  &  60.4 &  84.4  & Market &  51.4 & 72.4  \\
  DAAM~\cite{DBLP:conf/aaai/HuangPJLX20}    & AAAI2020 & Duke  & 67.8  & 86.4   & Market & 63.9  & 77.6  \\
  HGA~\cite{DBLP:conf/aaai/ZhangLLGDLJ21}   & AAAI2021 & Duke  & 70.3  & 89.5   & Market & 67.1  & 80.4  \\
  MMT~\cite{DBLP:conf/iclr/GeCL20}          & ICLR2020 & Duke  & 71.2  & 87.7   & Market & 65.1  & 78.0  \\
  JVTC+*~\cite{DBLP:conf/cvpr/ChenWLDB21}   & CVPR2021 & Duke  & 75.4  & 90.5   & Market & 67.6  & 81.9  \\
  MetaCam~\cite{DBLP:conf/cvpr/YangZLCLLS21}&CVPR2021  & Duke  & 76.5  & 90.1   & Market & 65.0  & 79.5  \\
  SpCL~\cite{DBLP:conf/nips/Ge0C0L20}       &NeurIPS2020& Duke & 76.7  & 90.3   & Market & 68.8  & 82.9  \\
  GCMT~\cite{DBLP:conf/ijcai/LiuZ21}        & IJCAI2021& Duke  & 77.1  & 90.6   & Market & 67.8  & 81.1  \\
  UNRN~\cite{DBLP:conf/aaai/ZhengLZZZ21}    & AAAI2021 & Duke  & 78.1  & 91.9   & Market & 69.1  & 82.0  \\
  GLT~\cite{DBLP:conf/cvpr/ZhengLH0LZ21}    & CVPR2021 & Duke  & 79.5  & 92.2   & Market & 69.2  & 82.0  \\
  OPLG~\cite{Zheng_2021_ICCV}               & ICCV2021 & Duke  & 80.0  & 91.5   & Market & 70.1  & 82.2  \\
  \hline
  \multicolumn{8}{l}{\textbf{Purely Unsupervised Learning}}    \\
  \hline
  BUC~\cite{DBLP:conf/aaai/LinD00019}  & AAAI2019 &  None &  38.3 & 66.2   & None  & 27.5  & 47.4   \\
  SSL~\cite{DBLP:conf/cvpr/LinXWY020}  & CVPR2020 &  None &  37.8 & 71.7   & None  & 28.6  & 52.5   \\
  MMCL~\cite{DBLP:conf/cvpr/WangZ20a}  & CVPR2020 &  None &  45.5 & 80.3   & None  & 40.2  & 65.2   \\
  HCT~\cite{DBLP:conf/cvpr/ZengNW020}  & CVPR2020 &  None &  56.4 & 80.0   & None  & 50.7  & 69.6   \\
  SpCL~\cite{DBLP:conf/nips/Ge0C0L20}  & NeurIPS2020 &None&  73.1 & 88.1   & None  & 65.3  & 81.2   \\
  IICS~\cite{DBLP:conf/cvpr/XuanZ21}   & CVPR2021 & None  & 72.1  & 88.8   & None  & 59.1  & 76.9 \\
  OPLG~\cite{Zheng_2021_ICCV}          & ICCV2021 & None  & 78.1  & 91.1   & None  & 65.6  & 79.8 \\
  CAP~\cite{DBLP:conf/aaai/0001LHG021} & AAAI2021 & None  & 79.2  & 91.4   & None  & 67.3  & 81.1 \\
  MGH~\cite{DBLP:conf/mm/WuWLT21}      & MM2021   & None  & 81.7  & 93.2   & None  & 70.2  & 83.7 \\
  ICE~\cite{DBLP:journals/corr/abs-2103-16364}   & ICCV2021 & None  & 82.3  & 93.8 & None  & 69.9  & 83.3 \\
  \hline
  \textbf{CA-UReID (Ours)}  &   -    & \textbf{None}  & \textbf{84.5} & \textbf{94.1}  & \textbf{None}  & \textbf{73.2} & \textbf{85.2} \\
  \hline
  \hline
  \end{tabular}
  \label{tab_sota_comparison}
  \vspace{-10pt}
\end{table}

\vspace{-5pt}
\subsection{Datasets and Implementation Details}
\vspace{-5pt}

\textbf{Datasets.} We use the mainstream ReID datasets to evaluate our proposed method: 
Market1501~\cite{DBLP:conf/iccv/ZhengSTWWT15} contains 32,668 images of 1,501 identities captured by 6 cameras, while DukeMTMC-reID~\cite{DBLP:conf/eccv/RistaniSZCT16} contains 36,411 images of 1,404 identities captured by 8 cameras.
As the common practice, Cumulative Matching Characteristic (CMC) and mean Average Precision (mAP) are used to measure the performance. 

\noindent \textbf{Implementation Details.}
We use ResNet-50 as backbone with the pre-trained weights on ImageNet. 
All images are resized to $256\times128$.
The data augmentation strategies used in training include random flipping, random cropping, and random erasing.
In each epoch, our method uses DBSCAN clustering with k-reciprocal Jaccard distance to generate pseudo-labels. 
The $eps$ in DBSCAN is set to 0.5/0.6 for Market1501/DukeMTMC-reID. 
Momentum factor $\mu$ is set to 0.2/0.1 on Market1501/DukeMTMC-reID.
In $\mathcal{L}_{cacc}$, the temperature factor $\tau_{cacc}$ is set to 0.07, and $J$ is set to 50.
The $\lambda_{1}$ and $\lambda_{2}$ in $\mathcal{L}_{total}$ is empirically set to 0.4 and 1.0,
The batch size is set to 64 with Adam optimizer. 
The learning rate is set to 0.00035 and multiplied by 0.1 per 20 epoch in training.

\vspace{-10pt}
\subsection{Comparison with the State-of-the-art}
\vspace{-5pt}

In experiments, we compare our method with state-of-the-art unsupervised ReID methods, including the UDA and USL methods.
The experimental results are shown in Tab.~\ref{tab_sota_comparison}.

We list state-of-the-art UDA methods in the top half of Tab.~\ref{tab_sota_comparison}.
The UDA leverages the annotation information of the source domain to make it easier to achieve better performance on the target dataset. 
Nevertheless, without any identity annotation, the performance of our method still clearly surpasses them, outperforming the latest OPLG method with +4.5\% mAP and +2.6\% Rank-1 on Market1501 and +3.1\% mAP and +3.0\% Rank-1 on DukeMTMC-reID.
As shown in the bottom half of Tab.~\ref{tab_sota_comparison}, our method also achieves better performance than SOTA methods under the purely unsupervised setting.
Compared with IICS~\cite{DBLP:conf/cvpr/XuanZ21}, which also considers the influence of camera style, our method has obvious advantages. 
IICS divides the training process into the intra-camera training stage and the inter-camera training stage, requiring additional AIBN to integrate the knowledge learned under different cameras and reduce the intra-camera variance. 
Our proposed camera separation structure can separate the feature map, which is more direct than IICS.
Compared with CAP~\cite{DBLP:conf/aaai/0001LHG021} and ICE~\cite{DBLP:journals/corr/abs-2103-16364} using camera-aware proxys, our method still achieve better performance on both Market1501 and DukeMTMC-reID. 
It also indicates that our designed center-based cross-camera loss has superiority with less sensitivity to noise samples. 

\vspace{-10pt}
\subsection{Ablation Study}
\vspace{-5pt}

\begin{table}[t]
  \centering
  \caption{Ablation studies of the designed components.}
  \fontsize{9}{9.5}\selectfont
  \setlength{\tabcolsep}{1.5mm}
  \begin{tabular}{c|l|cc|cc}
  \hline
  \hline
  \multicolumn{1}{c|}{\multirow{2}[0]{*}{Index}} & \multicolumn{1}{c|}{\multirow{2}[0]{*}{Methods}} & \multicolumn{2}{c|}{Market1501} & \multicolumn{2}{c}{Duke-reID} \\\cline{3-6}
      &  & mAP   & Rank1    & mAP   & Rank1 \\
  \hline
  1 & Baseline & 78.7  & 91.3  & 68.6  & 82.2 \\
  2 & Baseline~+~CSS & 80.7  & 91.9  & 69.5  & 82.4 \\
  3 & Baseline~+~$\mathcal{L}_{cacc}$ & 82.8  & 93.3  & 71.9  & 85.0 \\
  4 & Baseline~+~CSS~+~$\mathcal{L}_{cacc}$ & \textbf{84.5}  & \textbf{94.1}  & \textbf{73.2}  & \textbf{85.2} \\
  \hline
  5 & $\mathcal{L}_{casc}$ (with sample) & 83.5  & 93.4  & 72.0  & 84.0 \\
  6 & $\mathcal{L}_{cacc}$ (with center) & \textbf{84.5}  & \textbf{94.1}  & \textbf{73.2}  & \textbf{85.2} \\
  \hline
  \hline
  \end{tabular}
  \label{tab_ablation}
  \vspace{-5pt}
\end{table}

\begin{figure}[t]
  \centering
  \includegraphics[width=\linewidth]{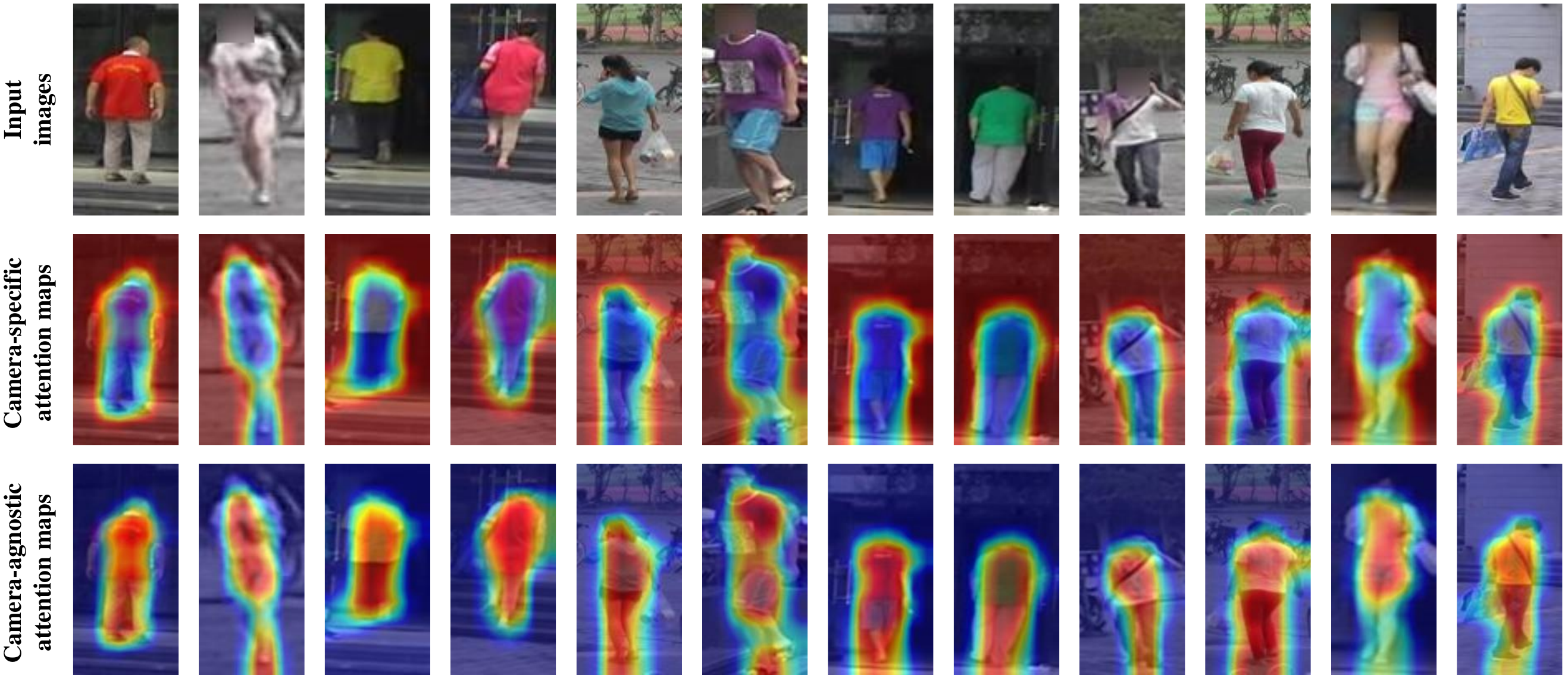}
  \vspace{-22pt}
  \caption{Visualization of attention map in the CSS module.}
  \vspace{-15pt}
  \label{fig_css_attention}
\end{figure}

In this section, we further discuss and analyze the effectiveness of key components, including the camera style separation module and the camera-aware contrastive center loss.

As shown in Tab.~\ref{tab_ablation}, the Baseline (Index-1) is the method without our proposed CSS branch and CACC loss.
Compared with the Baseline, the CSS module brings performance improvements on all two metrics (mAP, Rank1) are 2.0\%, 0.6\% on Market1501 and 0.9\%, 0.2\% on DukeMTMC-reID. 
The improvements of $\mathcal{L}_{cacc}$ (Index-3) over baseline are 4.1\%, 2.0\% on Market and 3.3\%, 2.8\% on Duke.
The best performance can be obtained by using both CSS and $\mathcal{L}_{cacc}$ (Index-4), with improvements of 5.8\%, 2.8\% on Market and 4.6\%, 3.0\% on Duke. 
It shows the good complementarity of CSS and $\mathcal{L}_{cacc}$ and they can jointly optimize the feature learning.

To observe the function of CSS module, we visualize the learned attention maps in the HA structure.
As shown in Fig.~\ref{fig_css_attention}, the camera-specific maps mainly focus on background, which is more related to cameras because of their fixed position.
In contrast, the complementary camera-agnostic attention maps focus on the main body of pedestrian, which can narrow the gap between cameras within the class.
In experiments, we also compare two different constraints: the camera-aware contrastive sample loss $\mathcal{L}_{casc}$ and camera-aware contrastive center loss $\mathcal{L}_{cacc}$, which act on samples and centers, respectively.
From Tab.~\ref{tab_ablation}, we can notice that the version of center-based loss achieves higher performance with stronger retrieval ability.
In general, by using two components together, our method can generate more discriminative embeddings in this person ReID task under USL settings.

\vspace{-10pt}
\section{Conclusion}
\vspace{-5pt}

In this paper, we propose a camera-aware style separation and contrastive learning method for purely unsupervised person ReID task, which can effectively separate camera style from the feature map by our proposed camera style separation module and camera-aware contrastive center loss. 
Experimental results demonstrate that our method shows superior performance against the existing state-of-the-art methods.

\clearpage
\bibliographystyle{IEEEbib}
\begin{spacing}{0.9}
\bibliography{icme2022_paper}
\end{spacing}

\end{document}